\documentclass[letterpaper, 10 pt,conference]{ieeeconf}
\IEEEoverridecommandlockouts                              

\usepackage{cite}
\usepackage{amsmath,amssymb,amsfonts}
\usepackage{graphicx}
\usepackage{textcomp}
\usepackage{xcolor}
\usepackage{algorithm}
\usepackage{algcompatible}
\usepackage{microtype}
\usepackage{array}
\usepackage{tikz,}
\usepackage{graphicx}
\usepackage{tabularx}
\usepackage{multirow}
\usepackage[utf8]{inputenc}
\usepackage[font=footnotesize]{subfig}
\usetikzlibrary{positioning}
\usetikzlibrary{arrows}
\IEEEoverridecommandlockouts   
\usepackage{caption}
\newcolumntype{P}[1]{>{\centering\arraybackslash}p{#1}}
\newcolumntype{M}[1]{>{\centering\arraybackslash}m{#1}}

\makeatletter
\newcommand{\removelatexerror}{\let\@latex@error\@gobble}
\makeatother

\begin{document}

\title{\LARGE \bf
ST-GIN: An Uncertainty Quantification Approach in Traffic Data Imputation with Spatio-temporal Graph Attention and Bidirectional Recurrent United Neural Networks
}

\author{Zepu Wang$^{1}$, Dingyi Zhuang$^{2}$, Yankai Li$^{1}$, Jinhua Zhao$^{2}$, Peng Sun$^{3^*}$, Shenhao Wang$^{4}$, Yulin Hu$^{5}$
\thanks{$^{1}$ Department of Computer and Information Science, University of Pennsylvania; {\tt\small zepu@seas.upenn.edu};
{\tt\small liyankai@seas.upenn.edu}}
\thanks{$^{2}$ Department of Civil and Environmental Engineering, Massachusetts Institute of Technology; {\tt\small dingyi@mit.edu}; {\tt\small jinhua@mit.edu}}%
\thanks{$^{3}$ Department of Natural and Applied Sciences, Duke Kunshan University; {\tt\small peng.sun568@duke.edu}}%
\thanks{$^{4}$ Department of Urban and Regional Planning, University of Florida; {\tt\small shenhaowang@ufl.edu}}%
\thanks{$^{5}$ Department of Information and Communication Engineering, Wuhan University; {\tt\small hu@isek.rwth-aachen.de}}%
\thanks{$^{*}$ Corresponding Author \newline
This work is supported by the National Natural Science Foundation of China (Grant No. 62250410368)}
}

\maketitle



\begin{abstract}
Traffic data serves as a fundamental component in both research and applications within intelligent transportation systems. However, real-world transportation data, collected from loop detectors or similar sources, often contains missing values (MVs), which can adversely impact associated applications and research. Instead of discarding this incomplete data, researchers have sought to recover these missing values through numerical statistics, tensor decomposition, and deep learning techniques. In this paper, we propose an innovative deep learning approach for imputing missing data. A graph attention architecture is employed to capture the spatial correlations present in traffic data, while a bidirectional neural network is utilized to learn temporal information. Experimental results indicate that our proposed method outperforms all other benchmark techniques, thus demonstrating its effectiveness.

\end{abstract}

\section{Introduction}
\label{intro}
As a crucial public resource responsible for ensuring effective communication among personnel and seamless circulation of materials, the transportation system's efficient and stable operation is pivotal to maintaining the smooth functioning of modern society~\cite{wang2022novel}. In this context, traffic data assumes a fundamental role in facilitating applications and research in the transportation domain. It is indispensable for both individuals seeking route planning solutions and researchers and governments involved in transportation management and control~\cite{wang2010parallel}.


 Notably, traffic data collected from loop detectors or other channels is frequently incomplete, owing to various reasons, which poses challenges for traffic analysis and other operations in practice~\cite{smith2003exploring}. In this regard, despite technological advancements, the issue of missing data remains a persistent challenge that is difficult to address. For instance, according to Chandra and colleagues, data collected by loop detectors on I-4 in Orlando, Florida had a missing rate of 15 percent~\cite{al2004new}. The ST data collection of the Georgia NaviGAtor system had an average rate of missed data ranging from 4 percent to 14 percent~\cite{ni2005markov}.

Traffic data can be categorized as spatial-temporal data, and it exhibits two critical characteristics~\cite{jiang2022deep}. Firstly, it demonstrates temporal dependence, implying the existence of non-linear temporal dependencies. For instance, traffic situations may vary dynamically, periodically, and regularly (e.g., during morning and evening rush hours), leading to changes in the correlations between different time steps. Another characteristic of traffic data is its spatial dependence, which implies the presence of dynamic spatial connections on complex networks. This means that the relationships between nodes in the road network can change over time, depending on various traffic situations. For instance, traffic congestion can have a negative effect on traffic upstream, but it may not affect traffic downstream as much.
 
In recent times, the advancements in deep learning have facilitated the integration of artificial intelligence into numerous real-world applications. In this research, we have proposed an innovative deep learning framework, namely ST-GIN (SpatioTemporal-Graph Imputation Network with Uncertainty Quantification), that combines graph attention neural networks and bidirectional recurrent united neural networks. We have leveraged uncertainty quantification to perform the task of data imputation.

The paper is structured as follows: Section~\ref{Literature} provides a comprehensive review and summary of previous studies on trajectory prediction. Section~\ref{Problem Statement} introduces the problem of missing data imputation using deep learning. The proposed methodology and the adopted loss function are described in Section~\ref{Proposed Method}. Section~\ref{EXPERIMENT} presents the experimental results that compare the performance of several models against our proposed approach. Finally, the contributions of this study are summarized in Section~\ref{Conclusion}.

\section{Literature Review}
\label{Literature}

\subsection{Imputation in traffic data}
To address incomplete traffic data, a basic strategy involves discarding entire rows containing MVs; however, this approach risks losing valuable information. A more effective strategy entails preprocessing the data by imputing MVs, i.e., inferring them from the known parts of the data~\cite{boquet2019missing, wang2023low}. Various imputation techniques are proposed to handle missing data in traffic datasets, with the primary categories being tensor decomposition-based and deep learning-based methods.

In decomposition-based methods, the core idea is to represent the original data in a more compact or low-rank form, then reconstruct the full data to estimate the missing values. Bayesian tensor decomposition methods are widely utilized for transforming inputs into low-rank factors and subsequently reconstructing the complete tensor~\cite{chen2021bayesian, chen2021low}.

Nonetheless, decomposition-based methods are constrained by the optimization norm and model assumptions. In contrast, deep learning methods have recently gained traction due to their powerful representation capabilities. By incorporating both spatial and temporal knowledge, the inductive properties of neural networks offer an effective means of imputing missing values in traffic datasets~\cite{wu2021inductive, wu2021spatial}.

\subsection{Deep Learning and graph neural networks}
Deep learning is a machine learning subfield that involves training neural networks with multiple layers to learn and represent complex patterns in data~\cite{lecun2015deep}. It has brought about significant advancements in many areas of research and industry. Notable deep learning architectures include Convolutional Neural Networks (CNNs) for image processing~\cite{gu2018recent}, Recurrent Neural Networks (RNNs) for sequential data~\cite{wang2022sfl}, and Generative Adversarial Networks (GANs) for generating realistic images and data~\cite{creswell2018generative}.

Graph Neural Networks (GNNs) have recently garnered considerable attention in the field of artificial intelligence and machine learning. GNNs are a type of deep learning model that can directly operate on graph-structured data. In transportation research, GNNs have gained popularity due to their capability of handling complex spatial data, such as traffic flow networks and urban transportation systems \cite{zhuang2022uncertainty,jiang2023uncertainty}.

\subsection{Uncertainty Quantification}

Research on uncertainty quantification has progressed in various deep learning fields. In this study, we focus on handling data uncertainty, which refers to the irreducible uncertainty inherent in the data generation process. For example, in linear regression, data uncertainty corresponds to the residuals, which are typically assumed to follow a normal distribution.

Data uncertainty can be characterized using parametric methods. Parametric methods involve models that parameterize a probabilistic distribution, which is commonly estimated through Bayesian methods or mean-variance estimation (MVE) \cite{wang2023uncertainty,zhuang2022uncertainty}. Despite their conceptual appeal, Bayesian methods often necessitate intensive computation, relying on sampling methods and variational inference \cite{Pearce2018, Khosravi2011}. In contrast, MVE minimizes the negative log-likelihood (NLL) loss based on a pre-specified distribution of the dependent variable \cite{Nix1994,khosravi2014optimized}. Although MVE is computationally efficient, it can yield misleading results if probabilistic distributions are misspecified.


\section{Problem Statement}
\label{Problem Statement}

Formally, a graph $G$ is defined as an ordered pair $(V,E)$, where $V$ is the set of vertices (or nodes) and $E$ is the set of edges. In the context of traffic data, the topology of a road network can be represented as a graph. The set of edges $E$ in the graph reflects the connections between road segments, while the set of vertices $V$ stores the traffic feature (e.g. speed and flow) of the road segments. Thus, the urban road network can be represented as a graph $G = (V,E)$, where the set of road segments $V=\left\{v_1, v_2, \cdots, v_N \right\}$ denotes has a size of $N$. Equivalently, the connectivity information $E$ between the road segments is stored in the adjacency matrix $A \in \mathbb{R}^{N \times N}$, where the entry $A_{ij}$ indicates the connectivity between the $i$th and $j$th road segments. In this paper, we adopt a well-known adjacency matrix construction technique from~\cite{li2017diffusion} that a value of 0 in this entry indicates no connection, while a non-negative value indicates the weight of the edge between the two vertices $v_i$ and $v_j$. We set the diagonal entries of $A$ to 1, as the weight matrix is given by $V_{ij} = exp(-\frac{dist(v_i, v_j)^2}{\sigma^2})$.

The matrix of features $X$, belonging to the set of real numbers $\mathbb{R}^{N \times T}$, corresponds to the traffic features $V$. Here, $N$ represents the number of sensors located between time $t$ and $t+T$, where $[t, t+1, \dots, t+T]$ represents a sequence of evenly spaced, continuous time instances. The $X_{ti}$ entry stands for the traffic information recorded for all road segments at time $ti$, while the $X_n$ entry captures the traffic information of sensor $n$ between time $t$ and $t+T$, where $n$ ranges from 1 to $N$.

During the operation of these $N$ sensors, certain situations may arise leading to the problem of missing data. Specifically, data may be partially occluded during the observed time period for a particular sensor $n$ due to data storage or transmission failure. In this paper, this issue is defined as \textit{random missing}.

On the other hand, there might be instances when some sensors experience system failure leading to the loss or occlusion of all data during the observed time period. In such scenarios, all values within $X_n$ are missing, and this paper defines this situation as \textit{non-random missing}.

Mathematically, $X_{t}$, $X_{t+1}$, $\dots$, and $X_{t+T}$ represent the univariate traffic data with real values and missing values of different time steps. Similarly, $\hat{X}_{t}$, $\hat{X}_{t+1}$, $\dots$, and $\hat{X}_{t+T}$ represent the imputed data by the specific function $f$.

\begin{equation}
\left[\hat{X}_{t}, \hat{X}_{t+1} \cdots, \hat{X}_{t+T}\right]=f\left(G ;\left(X_{t}, X_{t+1}, \cdots, X_{t+T}\right)\right).
\end{equation}

However, to view this reconstruction problem as a Bayesian perspective, we assume that all the traffic data is generated from an unknown Gaussian distribution:

\begin{equation}
    X_t \sim \mathcal{N}(\mu_t, \sigma_t^2).
\end{equation}

Hence, in the context of missing data, we would like to approximate ($\mu_t$, $\sigma_t^2$), ($\mu_{t+1}$, $\sigma_{t+1}^2$), $\dots$, ($\mu_{t+T}$, $\sigma_{t+T}^2$) based on the existing data.
\section{Proposed Method}
\label{Proposed Method}

In this section, we first introduce the concept of graph attention networks and bidirectional gated recurrent united networks. Then, we explain the proposed deep learning architecture.

\subsection{Graph Attention Networks}

Graph attention networks (GATs)~\cite{velickovic2017graph} have emerged as a powerful tool for modeling complex relationships between nodes in a graph. GATs employ attention mechanisms to selectively aggregate information from neighboring nodes, enabling them to capture both local and global patterns in the graph.

To achieve this, GATs first compute attention coefficients $\alpha_{ij}$ for each pair of neighboring nodes. Once the attention coefficients are computed, they are used to aggregate information from neighboring nodes. Specifically, the embedding for node $i$ is computed as a weighted sum of the embeddings of its neighbors:

\begin{equation}
h_i = \gamma \Biggl( \sum_{j \in \mathcal{N}(i)} \alpha_{ij} W x_j \Biggr).
\end{equation}

Here, $\mathcal{N}(i)$ represents the set of neighbors of node $i$, $W \in \mathbb{R}^{d \times f}$ is a weight matrix, $x_j$ represents the inputs of the layer, and $\gamma$ is an activation function.

The attention coefficients $\alpha_{ij}$ are computed as follows:

\begin{equation}
\alpha_{ij} = \frac{\exp(\text{LeakyReLU}(a^T [W x_i \| W x_j]))}{\sum_{k \in N(i)} \exp(\text{LeakyReLU}(a^T [W x_i \| W x_k]))},
\end{equation}

where $a \in \mathbb{R}^{2f}$ is a weight vector, $\|$ denotes concatenation, and $\text{LeakyReLU}$ is a leaky rectified linear unit activation function with negative slope $\alpha$. The attention coefficients $\alpha_{ij}$ are learned during training via backpropagation, allowing the model to adaptively focus on different parts of the graph as needed. In a traffic network, it is crucial to gather information from the surrounding nodes to effectively contribute to the imputation of missing data for a specific node \cite{wu2021inductive}.



\subsection{Bidirectional Recurrent Neural Networks}
Bidirectional Recurrent Neural Networks (Bi-RNNs) are a powerful neural network architecture that enables the flow of information in both the forward and backward directions through the recurrent layer~\cite{schuster1997bidirectional}. In particular, the bidirectional gated recurrent unit (BiGRU) is a variant of the recurrent neural network that can effectively process sequential data by analyzing it in both the forward and backward directions. Unlike the traditional gated recurrent unit (GRU)~\cite{cho2014learning}, the BiGRU incorporates an additional set of GRU cells that process the data in reverse. As a result, the BiGRU is capable of capturing both the past and future context of a sequence, making it a valuable tool in many applications.

The BiGRU can be represented mathematically as follows: Let $a_t$ be the input at time step $t$, and $h_t$ be the hidden state of the BiGRU at time step $t$. The final hidden state of the BiGRU at time step $t$ is obtained by concatenating the forward and backward hidden states:

\begin{equation}
\hat{h_t} = [h_t \| h'_t].
\end{equation}


This final hidden state contains information about both the past and future context of the input sequence.


In the context of data imputation, traditional time series forecasting problems only rely on previous time series data to predict future data. However, data imputation can also utilize future data to backwardly deduct previous time series data. As such, the BiGRU architecture is well-suited to address the data imputation problem, as it can leverage both forward and backward sequential temporal information to effectively fill in missing values.

\subsection{The general structure}
\label{structure}

\begin{figure}[!htbp]
	\centering
	\includegraphics[width=1\linewidth]
 {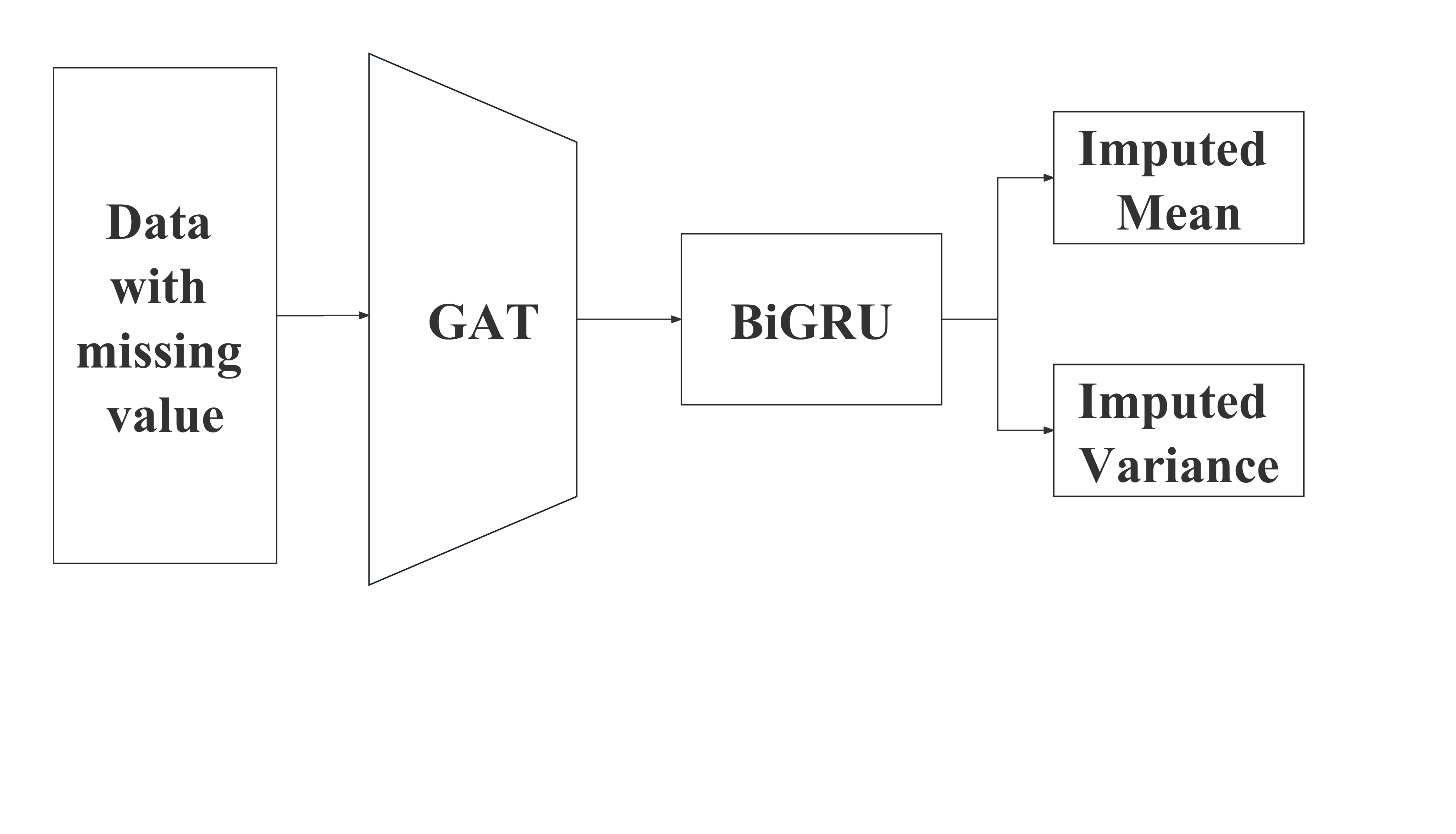}
	\caption{The general structure of ST-GIN}
	\label{fig: Auto-encoder}
\end{figure}

Our proposed approach for imputing missing data in traffic networks employs a two-step process, leveraging GATs an BiGRUs to address both the spatial and temporal dependencies in the data, shown in Figure \ref{fig: Auto-encoder}.

Firstly, we utilize a GAT layer to capture spatial dependencies within the traffic road network. This GAT architecture generates a spatial representation of the missing data by considering the relationships between neighboring nodes. The graph attention mechanism assigns different weights to neighboring nodes, emphasizing relevant connections and reducing the impact of distant or unrelated nodes. This helps in understanding localized patterns and road segment interactions, which are essential for accurate imputation.

For the second step, we employ a BiGRU layer to capture the temporal features of the traffic data. BiGRU layer processes the data in both forward and reverse directions, enabling the model to capture historical trends, real-time fluctuations, and future patterns in traffic flow. This dual processing enhances the imputation quality by grasping temporal context and dependencies, leading to more accurate predictions.

By integrating spatial and temporal information, we output the mean $\hat{\mu}$ and variance $\hat{\sigma}^{2}$ of the approximated missing data, providing imputation results and uncertainty quantification for the missing values based on Gaussian distributions.

\subsection{Loss function}
During the training process, we implement the training framework of variational autoencoder~\cite{kingma2013auto} and make some adjustments. To view the proposed method in another prospective, the architecture in ~\ref{structure} is identical to the Encoder component in the variational autoencoder (VAE) framework. During the training of variational autoencoder, the loss function is divdied into two parts,  a reconstruction loss and a regularization loss. The reconstruction loss measures how well the VAE can reconstruct the input data, while the regularization loss encourages the VAE to have a well-behaved latent space.

\subsubsection{reconstruction loss}

Instead of generating the traffic data randomly based on $\hat{\mu}$ and $\hat{\sigma}$, since the $\hat{\mu}$ has the highest probability to be generated, we directly assign the generated data to be $\hat{\mu}$. At the same time, since some of the values are missing, the loss function can only be calculated by non-missing values. Therefore, the reconstruction loss can be expressed as:

\begin{equation}
L_{Reconstruction} = \frac{1}{k}\sum_{i=1}^k(x_i-\hat{\mu_i})^2,
\end{equation}

here, $x_1,\ldots,x_k$ are non-missing real traffic data, and $(\hat{\mu}_1,\hat{\sigma}_1),\ldots,(\hat{\mu}_k,\hat{\sigma}_k)$ are the corresponding distributions generated by the neural networks.

\subsubsection{Regularization loss}
For the regularization loss, we would like to maximize the probability that the existing data are generated; hence, the negative log-likelihood (NLL) loss is adopted:
\begin{equation}
\begin{aligned}
L_{Regularization} &= \mathcal{L}(\hat{\mu}, \hat{\sigma} | x_1, x_2, ..., x_k) \\
&= -\ln\prod_{i=1}^k \frac{1}{\sqrt{2\pi\hat{\sigma_i}^2}}\exp\Big(-\frac{(x_i-\hat{\mu_i})^2}{2\hat{\sigma_i}^2}\Big) \\
&= -\sum_{i=1}^k\ln\Big(\frac{1}{\sqrt{2\pi\hat{\sigma_i}^2}}\Big)-\sum_{i=1}^k\Big(\frac{(x_i-\hat{\mu_i})^2}{2\hat{\sigma_i}^2}\Big) \\
&= -\frac{k}{2}\ln(2\pi\hat{\sigma_i}^2)-\frac{1}{2\hat{\sigma_i}^2}\sum_{i=1}^k(x_i-\hat{\mu_i})^2,
\end{aligned}
\end{equation}

here, $x_1,\ldots,x_k$ are non-missing real traffic data, and $(\hat{\mu}_1,\hat{\sigma}_1),\ldots,(\hat{\mu}_k,\hat{\sigma}_k)$ are the corresponding distributions generated by the neural networks.

\subsubsection{Combined loss}
In this research, we introduce a control hyperparameter $\lambda$ to control the influence of two sub loss functions to the total loss. For the experiment, we select $\lambda$ as 0.5 after trials and errors.
\begin{equation}
\mathcal{L} = \lambda \cdot L_{Reconstruction} + (1-\lambda) \cdot L_{Regularization}
\end{equation}

\section{Experiment}
\label{EXPERIMENT}

\subsection{Data Preparation}
In this research, we utilize the METR-LA dataset~\cite{li2017diffusion}, a publicly available dataset providing traffic speed and flow measurements from 207 detectors in 5-minute intervals on highways in Los Angeles County. The data is collected using inductive loop detectors installed on highways. The collected data is then aggregated and processed to generate traffic flow information, such as average speed and traffic volume, for different segments of the road network. The chosen time period is from March 1, 2012, to March 30, 2012.

To simulate missing values in the dataset, we use two scenarios: random missing and non-random missing. For the random missing case, we employ the traffic speed data and randomly select different portions (0.1, 0.2, 0.3, 0.4, 0.5, 0.6, 0.7) of traffic data inputs, setting them as MVs, represented by zeroes. This scenario simulates random temporary sensor failures or random data storage problems. For the non-random missing scenario, we utilize traffic flow data and randomly select different portions (0.1, 0.2, 0.3, 0.4) of sensors, setting all values for these sensors to zeroes. This scenario aims to simulate long-term sensor or system malfunctions.

The accuracy of an imputation model can be evaluated by comparing the predicted traffic mean to the actual observed value. Notice that traffic flow data are non-negative, we transform our negative mean values to zero. Consequently, we employ two popular metrics, Mean Absolute Error (MAE) and Mean Square Error (MSE):

\begin{equation}
MAE = \frac{1}{n} \sum_{i=1}^n \left| y_i - \hat{y_i} \right|,
\end{equation}
\begin{equation}
MSE = {\frac{1}{n}\sum_{i=1}^n(y_i-\hat{y_i})^2}.
\end{equation}

\subsection{Baseline Methods}
Here, we introduce the baseline methods that we use in this experiment.
\begin{itemize}

\item[$\bullet$]Average: For missing values in a specific day, fill them in using the average of non-missing values during the same time period.
\item[$\bullet$]Mean: For missing values in a specific day, fill them with the mean value of the same sensor in this specific day.
\item[$\bullet$]Singular Value Decomposition (SVD)~\cite{kurucz2007methods}:  a matrix factorization technique. Fill in missing values by estimating them based on the relationships between other variables in the dataset. 
\item[$\bullet$]Temporal Regularized Matrix Factorization (TRMF)~\cite{yu2016temporal}: an effective tool for imputing missing data within a given multivariate time series and forecasting time series with missing values. 
\item[$\bullet$]Bidirectional Gated Recurrent United neural networks (BiGRU): BiGRU is proficient in dealing with time series problems. 
\item[$\bullet$]Graph Convolutional Neural Networks (GCN)~\cite{kipf2016semi}: A neural network structure that is efficient in dealing with graph information.
\end{itemize}

However, TRMF, SVD, Mean and Average cannot deal with the non-random missing situation. Hence, all baseline methods are tested in random missing data. BiGRU and GCN are tested in non-random missing data. 

\subsection{Result and Analysis}
\subsubsection{Random Missing}
\begin{table*}[htbp]
\renewcommand{\arraystretch}{2.5}
\centering
\scriptsize
\caption{MSE and MAE for random missing in traffic speed data (miles per hour)}
\begin{tabular}{lcccccccc}
    \hline
\textbf{Model} & \textbf{0.1 (MSE/MAE)} & \textbf{0.2 (MSE/MAE)} & \textbf{0.3 (MSE/MAE)} & \textbf{0.4 (MSE/MAE)} & \textbf{0.5 (MSE/MAE)} & \textbf{0.6 (MSE/MAE)} & \textbf{0.7 (MSE/MAE)}\\ 
    \hline
Average & 0.1537 / 0.3198 & 0.1498 / 0.3167 & 0.1480 / 0.3140 & 0.1578 / 0.3244 & 0.1704 / 0.3368 & 0.1901 / 0.3571 & 0.2175 / 0.3843\\ 
Mean & 0.0827 / 0.2491 & 0.0827 / 0.2489 & 0.0826 / 0.2490 & 0.0827 / 0.2490 & 0.0827 / 0.2492 & 0.0828 / 0.2492 & 0.0828 / 0.2493\\ 
SVD & 0.00446 / 0.0550 & 0.0153 / 0.1049 & 0.0325 / 0.1542 & 0.0559 / 0.2032 & 0.0865 / 0.25350 & 0.1230 / 0.3027 & 0.1659 / 0.3521\\
TRMF & 0.0065 / 0.0701 & 0.0082 / 0.0789 & 0.0108 / 0.0900 & 0.0148 / 0.1050 & 0.0217 / 0.1266 & 0.0334 / 0.1583 & 0.0596 / 0.2112\\ 
BiGRU & 0.0022 / 0.0429 & 0.0089 / 0.0844 & 0.0210 / 0.1294 & 0.0398 / 0.1794 & 0.0695 / 0.2357 & 0.1024 / 0.2857 & 0.1509 / 0.3489 \\
GCN & 0.0325 / 0.1465 & 0.0432 / 0.1759 & 0.0589 / 0.2071 & 0.0810 / 0.2439 & 0.1126 / 0.2873 & 0.14263 / 0.3231 & 0.1852 / 0.3696\\
ST-GIN & \textbf{0.0011}  / \textbf{0.0300} & \textbf{0.0037} / \textbf{0.0538} & \textbf{0.0038} / \textbf{0.0522} & \textbf{0.0137} / \textbf{0.0957} & \textbf{0.0128} / \textbf{0.0537} & \textbf{0.0374}  / \textbf{0.1757} & \textbf{0.0450} / \textbf{0.1684} \\
\hline
\label{Table 1}
\end{tabular}
\end{table*}

Table ~\ref{Table 1} presents the MSE and MAE for all the random missing cases. The results demonstrate that, except for the Mean method, the imputation accuracy decreases as the amount of missing data increases. However, our proposed ST-GIN method consistently achieves the highest accuracy compared to other baseline methods for a given portion of missing data.

As previously discussed, traffic data exhibits spatial and temporal dependencies, and effectively capturing the spatial-temporal correlation of the existing information is crucial for successful data imputation. The Mean and Average methods do not utilize the complex road network structures, resulting in their simple imputers being unable to fully recover traffic information.

SVD and TRMF strive to capture the spatial and temporal information by identifying the relationships between rows and columns of a large data matrix. Nevertheless, without considering the prior knowledge $G$ of the graph, these methods cannot effectively utilize the spatial information, leading to limited performance.

Among the three deep learning methods, GCN focuses solely on the spatial features of the traffic data, while BiGRU captures only the temporal correlations of the speed data. In comparison, ST-GIN, which analyzes both spatial and temporal dependencies, consistently outperforms GCN and BiGRU in terms of imputation accuracy.

\subsubsection{Non-random Missing}

Table \ref{Table 2} summarizes the MSE and MAE errors for all the non-random missing cases. As with the random cases, imputation becomes more challenging as the missing values become more consecutive. Many existing imputation methods are not suitable for handling consecutive missing data, either spatially or temporally\cite{wu2021spatial}. When compared to the other two deep learning baselines, ST-GIN generally exhibits better performance, except for the MAE value in the 0.4 missing case.


\begin{figure}[!htbp]
	\centering
	\includegraphics[width=1.1\linewidth]
 {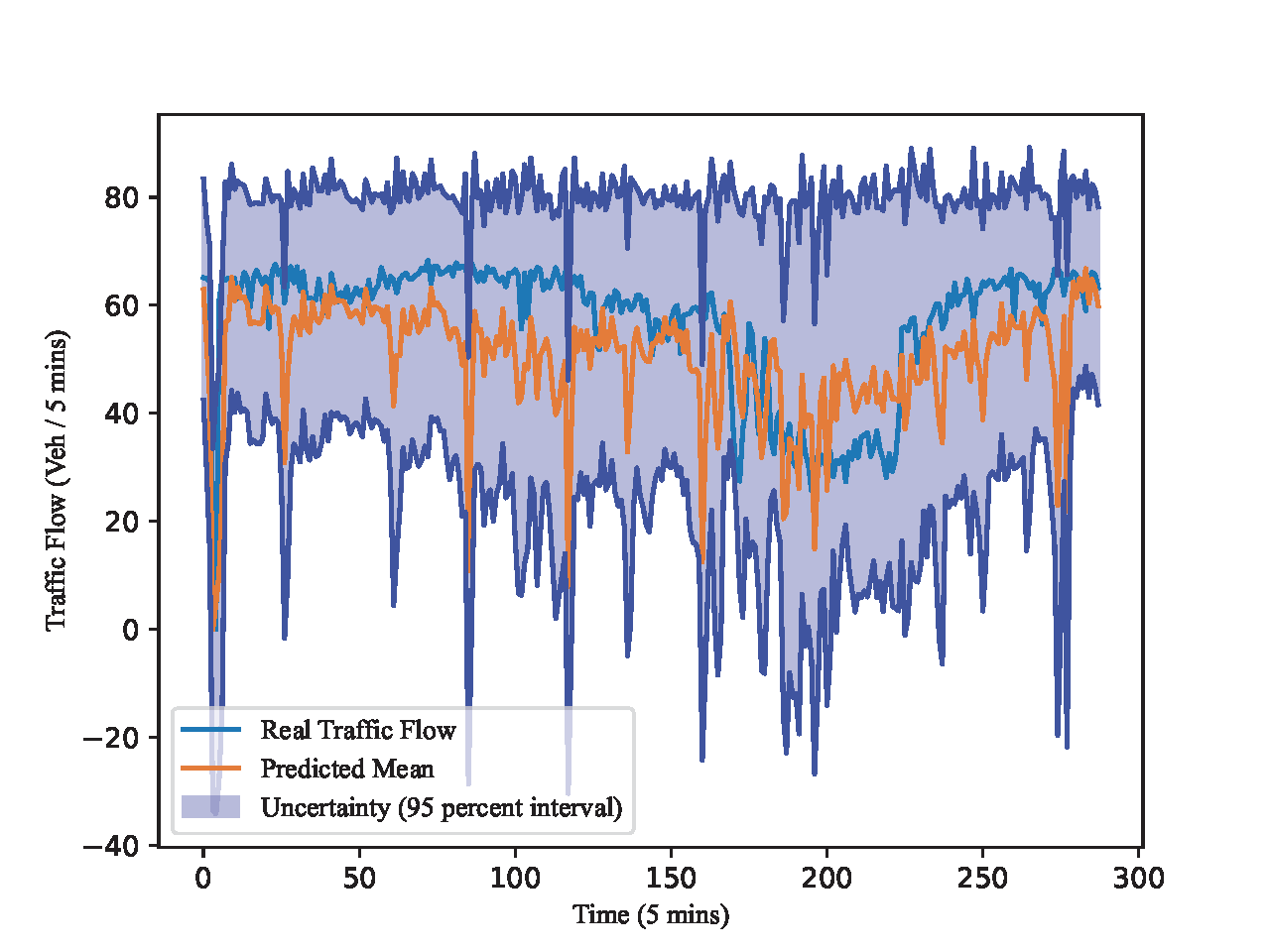}
	\caption{Uncertainty Quantification of 0.1 Non-random Missing Data}
	\label{fig: uncertainty}
\end{figure}

\begin{table}[htbp]
\renewcommand{\arraystretch}{1.8}
\centering
\caption{MSE and MAE in non-random missing traffic flow data (Veh per 5 mins)}
\begin{tabular}{llcccccc}
\hline
 \textbf{Metrics} & \textbf{Model} & \textbf{0.1} & \textbf{0.2} & \textbf{0.3} & \textbf{0.4} \\
\hline
\multirow{3}{*}{MSE} & BiGRU & 242.8  & 245.6 & 265.9 & 285.27 \\ 
& GCN & 252.6 & 286.5 & 322.6 & 375.9\\
& ST-GIN & \textbf{201.5} & \textbf{242.8} &\textbf{250.7} & \textbf{278.1}\\

\hline 

\multirow{3}{*}{MAE} & BiGRU & 10.72 & 11.62 & 11.79 & \textbf{12.75}\\
& GCN & 12.47 & 13.81 & 14.63 & 15.68\\
& ST-GIN     & \textbf{10.20} & \textbf{11.37} & \textbf{11.46} & 12.76\\
\hline
\label{Table 2}
\end{tabular}
\end{table}

Figure ~\ref{fig: uncertainty} plot a one-day value, imputed mean, and a 0.05 confidential interval of a random sensor when 10 percent of sensor data is missing. The imputed mean generally capture the trend of the real traffic flow data. However, due to the stochastic nonlinear nature of traffic flow, it is difficult to impute exact value of the traffic data, especially when the all values of the same data are missing. However, for all real values fall within in the 95 percent interval.

\section{Conclusion}
\label{Conclusion}

In this paper, we introduce a novel deep learning framework, ST-GIN, which effectively addresses the issue of missing data in traffic datasets. This framework leverages graph attention layers to capture the spatial relationships among traffic tensors while utilizing bidirectional gated recurrent neural networks to learn the temporal correlations of traffic data. Experimental results indicate that our method demonstrates superior performance when compared to numerous benchmark techniques for imputing missing speed data in both random and non-random missing scenarios, as exemplified by the METR-LA dataset.

Several potential avenues can be explored to further enhance and expand upon this research. One such direction includes employing a wider range of data to evaluate the adaptability of our model across various scenarios. This is particularly relevant for urban road networks, which are characterized by higher short-term variations due to uncertain road conditions and fluctuating traffic patterns. Investigating the model's performance in such complex environments will provide valuable insights into its applicability and robustness.

Additionally, integrating advanced deep learning frameworks, such as attention-based models and transformers, could further improve the imputation accuracy of our method. Last but not least, more external features, such as special weather conditions and traffic accidents, might be analyzed in the future.

\bibliographystyle{IEEEtran}
\bibliography{IEEEabrv,ref}
\end{document}